%% file: main.tex
\documentclass[letterpaper]{article} % DO NOT CHANGE THIS
\usepackage{aaai25}  % DO NOT CHANGE THIS
\usepackage{times}  % DO NOT CHANGE THIS
\usepackage{helvet}  % DO NOT CHANGE THIS
\usepackage{courier}  % DO NOT CHANGE THIS
\usepackage[hyphens]{url}  % DO NOT CHANGE THIS
\usepackage{graphicx} % DO NOT CHANGE THIS
\usepackage{natbib}  % DO NOT CHANGE THIS AND DO NOT ADD ANY OPTIONS TO IT
\usepackage{caption} % DO NOT CHANGE THIS AND DO NOT ADD ANY OPTIONS TO IT
\usepackage{algorithm}
\usepackage{algorithmic}
\usepackage{booktabs}

\usepackage{fancyvrb}
\newcommand\userinput[1]{\textbf{#1}} 
\usepackage{amsthm}
\newtheorem{definition}{Definition}
\usepackage{subcaption}
\usepackage{makecell}
\usepackage{newfloat}
\usepackage{listings}
\DeclareCaptionStyle{ruled}{labelfont=normalfont,labelsep=colon,strut=off} % DO NOT CHANGE THIS
\floatstyle{ruled}
\newfloat{listing}{tb}{lst}{}
\floatname{listing}{Listing}
\title{HDDLGym: A Tool for Studying Multi-Agent Hierarchical Problems Defined in HDDL with OpenAI Gym}
\author {
    Ngoc La\textsuperscript{\rm 1},
    Ruaridh Mon-Williams\textsuperscript{\rm 2},
    {\normalfont and} Julie A. Shah\textsuperscript{\rm 1}
}
\affiliations {
    \textsuperscript{\rm 1}MIT\\
    \textsuperscript{\rm 2}University of Edinburgh\\
    ntmla@mit.edu, ruaridh.mw@ed.ac.uk, julie.a.shah@csail.mit.edu
}

\usepackage{bibentry}
\usepackage{subcaption}
\begin{document}

\maketitle

\begin{abstract}
% In recent years, reinforcement learning (RL) methods have been widely tested using tools like OpenAI Gym, though many tasks in these environments could also benefit from hierarchical planning. However, there is no tool that enables seamless integration of hierarchical planning with RL. Hierarchical Domain Definition Language (HDDL), used in classical planning, introduces a structured approach well-suited for model-based RL to address this gap. To bridge this integration, we introduce HDDLGym, a Python-based tool that automatically generates OpenAI Gym environments from HDDL domains and problems. HDDLGym serves as a link between RL and hierarchical planning, supporting multi-agent scenarios and enabling collaborative planning among agents. This paper provides an overview of HDDLGym’s design and implementation, highlighting the challenges and design choices involved in integrating HDDL with the Gym interface and applying RL policies to support hierarchical planning. We also provide detailed instructions and demonstrations for using the HDDLGym framework, including how to work with existing HDDL domains and problems from International Planning Competitions, exemplified by the Transport domain. Additionally, we offer guidance on creating new HDDL domains for multi-agent scenarios and demonstrate the practical use of HDDLGym in the Overcooked domain. By leveraging the advantages of HDDL and Gym, HDDLGym aims to be a valuable tool for studying RL in hierarchical planning, particularly in multi-agent contexts.

In recent years, reinforcement learning (RL) methods have been widely tested using tools like OpenAI Gym, though many tasks in these environments could also benefit from hierarchical planning. However, there is a lack of a tool that enables seamless integration of hierarchical planning with RL. Hierarchical Domain Definition Language (HDDL), used in classical planning, introduces a structured approach well-suited for model-based RL to address this gap. To bridge this integration, we introduce HDDLGym, a Python-based tool that automatically generates OpenAI Gym environments from HDDL domains and problems. HDDLGym serves as a link between RL and hierarchical planning, supporting multi-agent scenarios and enabling collaborative planning among agents. This paper provides an overview of HDDLGym’s design and implementation, highlighting the challenges and design choices involved in integrating HDDL with the Gym interface, and applying RL policies to support hierarchical planning. We also provide detailed instructions and demonstrations for using the HDDLGym framework, including how to work with existing HDDL domains and problems from International Planning Competitions, exemplified by the Transport domain. Additionally, we offer guidance on creating new HDDL domains for multi-agent scenarios and demonstrate the practical use of HDDLGym in the Overcooked domain. By leveraging the advantages of HDDL and Gym, HDDLGym aims to be a valuable tool for studying RL in hierarchical planning, particularly in multi-agent contexts.
\end{abstract}

\begin{links}
    % \link{Code\footnote{The code link is included in the zip file submitted as supplementary materials.}}{TBD}
    \link{Code}{https://github.com/HDDLGym/HDDLGym}
    % \link{Datasets}{https://aaai.org/example/datasets}
    % \link{Extended version}{https://aaai.org/example/extended-version}
\end{links}

\input{sections/1_introduction}

\input{sections/2_background}

\input{sections/3_related_works}
\input{sections/4_formal_framework}
\input{sections/5_hddlgym_framework}
\input{sections/6_applications}
\input{sections/7_discussion_future_works_conclusions}

% \clearpage
\section*{Acknowledgments}
We gratefully acknowledge the financial support of the Office of Naval Research (ONR), under grant N000142312883, and sincerely thank Pulkit Verma for his valuable insights and feedback on the project.
% \small
% \clearpage
\bibliography{main}
\end{document}

%% file: sections/1_introduction.tex
\section{Introduction}

Hierarchical planning is essential for addressing complex, long-horizon planning problems by decomposing them into smaller, manageable subproblems. In reinforcement learning (RL), hierarchical strategies can guide exploration along specific pathways, potentially enhancing learning efficiency. 
% Hierarchical strategies can also introduce biases in reinforcement learning (RL) by directing exploration along specific pathways. 
However, implementing RL policies within hierarchical frameworks often requires custom modifications to the original environments to incorporate high-level actions \cite{wu_too_2021,liu2017learning,macro-action}. For example, in a Bayesian inference study using the Overcooked game, subtasks are integrated as high-level actions through specific rules embedded in the system codebase \cite{wu_too_2021}. Similarly, several RL studies use author-defined high-level actions, or macro-actions, to organize complex tasks \cite{liu2017learning,macro-action}. While these studies highlight the benefits of hierarchical approaches in complex 
% sequential decision-making
scenarios, the additional programming required to integrate hierarchical layers can make it challenging for external users to modify or implement alternative high-level strategies. This limitation reduces users' flexibility to implement diverse hierarchical strategies tailored to their specific requirements.

% - Bringing the benefits of using HDDL in defining domains and problems: can make use well-documented domains and problems from IPC-HTN, HDDL's intuitive and flexible design provides the convenience for users to easily define or modify problem-solving approaches to suit their needs, no limit on number of layers of action hierarchy provide flexibility in defining and solving the problems. 

% Hierarchical Domain Definition Language (HDDL), an extension of  PDDL developed by Holler et al. \cite{holler2020hddl}, offers a standardized language for hierarchical planning systems. One of its key advantages is the availability of well-documented domains and problems from sources like the International Planning Competitions' hierarchical task network tracks (IPC-HTN) \cite{IPC_HTN_tracks}. As another advantage, the HDDL's intuitive and flexible design provides the convenience for users to easily define or modify problem-solving approaches to suit their needs.
% To utilize the benefits of both HDDL and Gym, in this work, we propose HDDLGym as a Python tool to automatically generates Gym environment from HDDL domain and problem files in order to study RL in hierarchical planning problem.

The Hierarchical Domain Definition Language (HDDL) \cite{holler2020hddl} is an extension of Planning Domain Definition Language (PDDL) \cite{pddl} that incorporates hierarchical task networks (HTN) \cite{HTN}. HDDL provides a standardized language for hierarchical planning systems and is supported by extensive documentation as well as a variety of domains and problems. Many of these resources are sourced from the hierarchical task network tracks of the International Planning Competitions (IPC-HTN) \cite{IPC_HTN_tracks}. HDDL’s intuitive and flexible design also allows users to define or modify problem-solving approaches by adjusting the hierarchical task networks to suit their specific needs. To leverage HDDL’s strengths in studying RL within hierarchical planning problems, we present HDDLGym — a framework that integrates HDDL with OpenAI Gym \cite{brockman_openai_2016}, a standardized RL interface. HDDLGym is a Python-based tool that automatically generates Gym environments from HDDL domain and problem files.

% Multi-agent contexts are another significant topic in automatic planning in general, and hierarchical planning research in particular. HDDL is not inherently designed to manage multi-agent systems. The multi-agent feature has been studied in planning formalism, such as PDDL cite{mulitagent-PDDL}, hierarchical task network cite{multiagent-HTN}. However, to make use of well-documented and benchmarking HDDL domains and problem from IPC-HTN, our HDDLGym is designed to work with the original HDDL. To handle multi-agent contexts, we offer a protocol for defining HDDL domains and problems or suggest minor modifications to existing HDDL files to incorporate multi-agent features. 

Multi-agent contexts are a key area in automated planning and hierarchical planning research.
% Many planning scenarios involve multiple agents, and such problems have been explored in the context of planning~\cite{kovacs2012multi}.
While HDDL is not inherently designed for multi-agent systems, multi-agent features have been explored in planning formalisms like MA-PDDL \cite{kovacs2012multi} and MA-HTN \cite{cardoso2017multi}. However, to utilize the extensive, well-documented HDDL domains and problems from IPC-HTN, HDDLGym is designed to work closely with the HDDL defined by \citet{holler2020hddl}. 
% To support multi-agent contexts, we introduce a new protocol for defining HDDL domains and problems, which includes making minor modifications to existing HDDL files.
We introduce a new protocol for extending HDDL domains and problems to support HDDLGym with multi-agent features. This includes making minor modifications to existing HDDL files from IPC-HTN.

% This paper makes three key contributions. First, we introduce HDDLGym, a novel framework that automatically bridges reinforcement learning and hierarchical planning by automatically generating Gym environments from HDDL domains and problems. Second, we provide a protocol for modifying HDDL domains to support multi-agent configurations within HDDLGym, thereby extending hierarchical planning techniques to complex multi-agent environments. Third, we detail HDDLGym’s design and usage, demonstrating its effectiveness with examples from the Transport domain (IPC-HTN) and the Overcooked environment.

\begin{figure*}[t!]
 \begin{subfigure}[t]{0.5\textwidth}
    \centering
    \includegraphics[width=\textwidth]{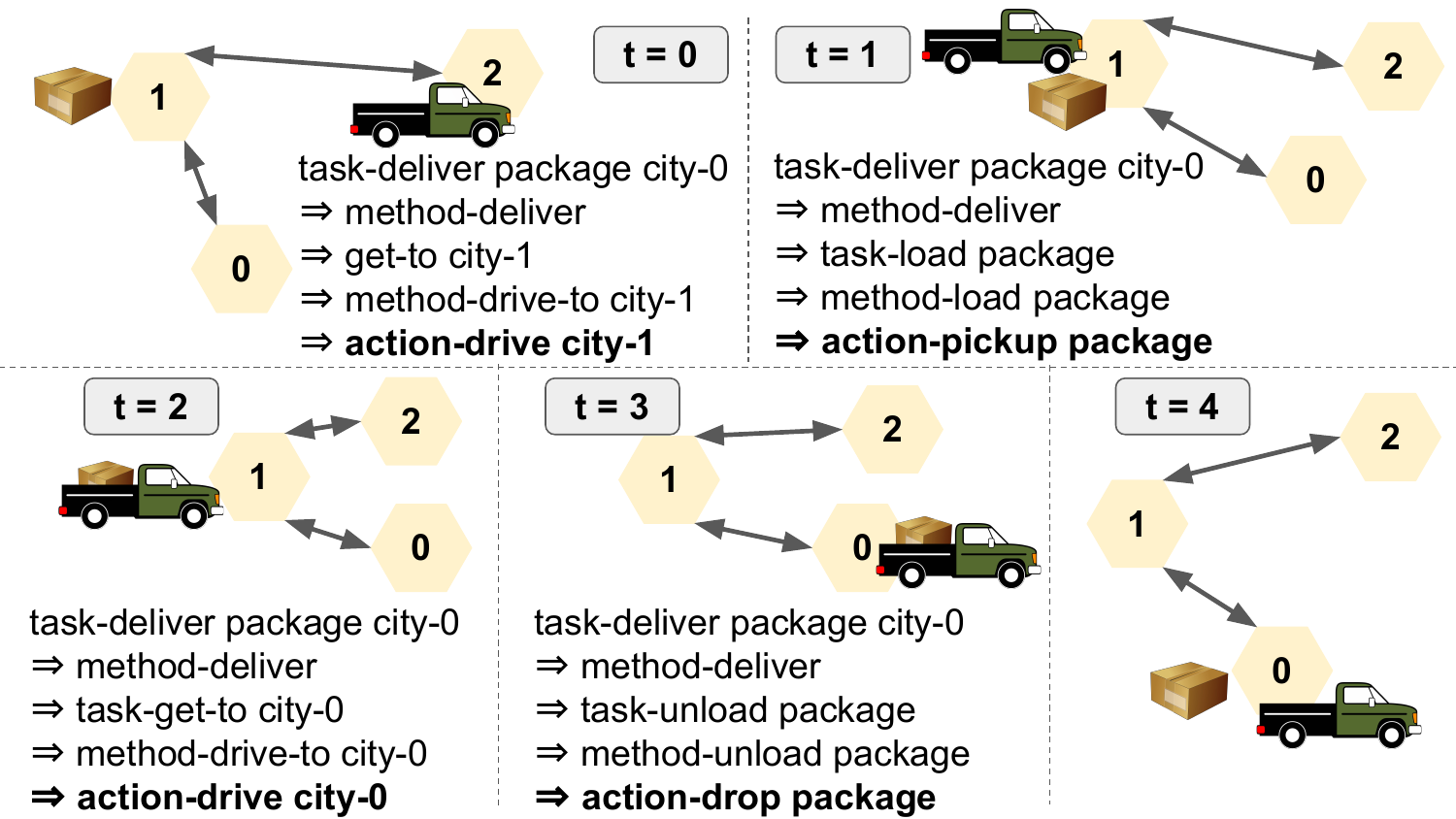} % Adjust width as needed
    \caption{Transport scenario}
    \label{fig:transport}
    \end{subfigure}
~
\begin{subfigure}[t]{0.5\textwidth}
    \centering
\includegraphics[width=0.94\textwidth,page=2]{images/overcooked_demo.pdf} % Adjust width as needed
    \caption{Overcooked scenario}
    \label{fig:overcooked}
\end{subfigure}
\caption{Examples of the Transport and Overcooked environments in HDDLGym}
\end{figure*}

% \noindent
\paragraph{Main contributions}
This paper makes the following three key contributions:
\begin{itemize} 
    \item We introduce HDDLGym, a novel framework that automatically bridges reinforcement learning and hierarchical planning by automatically generating Gym environments from HDDL domains and problems. 
    \item We provide a protocol for modifying HDDL domains to support multi-agent configurations within HDDLGym, thereby extending hierarchical planning techniques to complex multi-agent environments. 
    \item We detail HDDLGym’s design and usage, demonstrating its effectiveness with examples from the Transport domain (in IPC-HTN) and the Overcooked environment (as shown in Figures \ref{fig:transport} and \ref{fig:overcooked}, respectively).
\end{itemize}

\paragraph{Core features}
HDDLGym offers five core features, which are detailed throughout the paper.
\begin{itemize}
    \item Support verifying and adapting HDDL files to the tool (Section \ref{sec:formal});
    \item Support multi-agent collaborations (Sections \ref{subsec:planner} and \ref{sec:applications});
    \item Support centralized and decentralized planning (Section \ref{subsec:planner});
    \item Support modifying design of RL policy in hierarchical planning (Sections \ref{subsec:observation} and \ref{subsec:policy});
    \item Support training and deploying trained RL policies with various evaluation metrics and visualization (Sections \ref{sec:applications}).
\end{itemize}

The remainder of this paper is organized as follows. 
% The \textit{Background and Related Work} section 
Section \ref{sec:background}
provides background information on HDDL and OpenAI Gym, the two foundational frameworks on which our system is built. 
Section \ref{sec:relatedworks} discusses relevant prior work and thus highlights our contributions to the field.
Section \ref{sec:formal} then introduces the formal framework of HDDLGym, detailing how HDDL is modified to align with the agent-centric design of this tool.
Section \ref{sec:framework} covers the design and implementation details of HDDLGym.  
Following this, section \ref{sec:applications} demonstrates the use of HDDLGym with examples from the Transport domain, representing domains from IPC-HTN, and Overcooked, representing customized environments. 
Section \ref{sec:discussion} discusses the key benefits and current limitations of the HDDLGym tool, along with future developments to address these limitations and expand its applications within artificial intelligence research. Finally, Section \ref{sec:conclusion} concludes the paper.

%% file: sections/2_background.tex
\section{Background}
\label{sec:background}

\subsection{HDDL}
\label{subsec:HDDL}
% \subsubsection{HDDL}
% [Describe]
% An extension of PDDL \cite{pddl}, HDDL \cite{holler2020hddl} utilizes predicates to represent the current state of the environment.
HDDL \cite{holler2020hddl} is an extension of PDDL \cite{pddl}.
 \citet{holler2020hddl} define the domain and problem as follows.

\paragraph{Definition of Planning Domain:} A planning domain $D$ is a tuple $(L,T_P, T_C, M)$ defined as follows.
\begin{itemize}
    \item $L$ is the underlying predicate logic.
    \item $T_P$ and $T_C$ are finite sets of primitive and compound tasks, respectively.
    \item $M$ is a finite set of decomposition methods with compound tasks from $T_C$ and task networks over the set $T_P \cup T_C$.
\end{itemize}

\paragraph{Definition of Planning Problem:} A planning problem $\mathcal{P}$ is a tuple $(D, s_I, tn_I, g)$, where:
\begin{itemize}
    \item $s_I \in S$ is the initial state, a ground conjunction of positive literals over the predicates assuming the closed world assumption.
    \item $tn_I$ is the initial task network that may not necessarily be
grounded.
    \item $g$ is the goal description, being a first-order formula over the predicates (not necessarily ground).
\end{itemize}

In other words, beyond the action definition in PDDL, which establishes the rules of interaction with the environment, HDDL introduces two additional operators:  \textit{task} and \textit{method}. 
In HDDL, a \textit{task} represents a high-level action, while a \textit{method} is a strategy to accomplish a task. 
Multiple methods can exist to perform a single task. Essentially, a method is a task network that decomposes a high-level task into a partially or totally ordered list of tasks and actions.  

In HDDL, a task is defined with its parameters, and a method is defined with parameters, the associated task, preconditions, a list of subtasks with their ordering or a list of ordered subtasks. Examples of task and method definitions from the original HDDL work \cite{holler2020hddl} are:
\begin{quote}
\begin{scriptsize}
\begin{Verbatim}[numbers=left]
(:task get-to :parameters (?l - location))
(:method m-drive-to-via
    :parameters (?li ?ld - location)
    :task (get-to ?ld)
    :precondition ()
    :subtasks (and
        (t1 (get-to ?li))
        (t2 (drive ?li ?ld)))
    :ordering (and
        (t1 < t2)))
\end{Verbatim}
\end{scriptsize}
\end{quote}

In HDDL, state-based goal definition is optional. Goals are instead defined as a list of goal tasks in the HDDL problem file. An example of the goal in a transport problem is as follows.
\begin{quote}
\begin{scriptsize}
\begin{Verbatim}[numbers=left]
(:htn
    :tasks (and
        (deliver package-0 city-loc-0)
        (deliver package-1 city-loc-2))
    :ordering ())
\end{Verbatim}
\end{scriptsize}
\end{quote}

More details about HDDL domain and problem files can be found in \citet{holler2020hddl}. In addition to the original format of HDDL, some modifications are required to make the HDDL domains and problems work smoothly with HDDLGym. Details of the modifications are in Section~\ref{sec:formal}.

% [HDDLGym main parser: converting HDDL domain and problem files into a dictionary of essential components - list all keys]
% In order to convert the HDDL domain and problem files into Gym environment, HDDLGym reads and converts HDDL files into a Python dictionary of essential components in the \texttt{main\_parser.py}. Then, it constructs Gym environment based on the information from the environment dictionary. Therefore, users can easily edit the HDDL files to modify the environments' features, such as interaction rules, hierarchical strategies, and problem settings.
 
\subsection{OpenAI Gym}
% OpenAI Gym \cite{brockman_openai_2016} has become a widely adopted toolkit that offers a standardized interface for benchmarking and developing reinforcement learning (RL) algorithms. Its consistent application programming interface (API) includes key methods for environment initialization, resetting, and interaction, allowing researchers to focus on advancing RL algorithms without handling environment-specific implementation details. The toolkit includes a diverse set of environments, ranging from simple control tasks to complex simulations like Atari games, providing a common platform that enhances reproducibility and enables direct comparisons across different RL methodologies. Therefore, integrating OpenAI Gym with HDDL enables the development of a unified framework for designing and evaluating hierarchical RL approaches, combining the adaptive learning strengths of RL with the structured decision-making of hierarchical planning.
OpenAI Gym \cite{brockman_openai_2016} is a widely adopted toolkit that provides a standardized interface for benchmarking and developing reinforcement learning (RL) algorithms. Its consistent API includes methods for environment initialization, resetting, and interaction, allowing researchers to focus on RL algorithm development without dealing with environment-specific details. With a diverse range of environments, from simple tasks to complex simulations like Atari games, Gym enhances reproducibility and enables direct comparisons across RL methodologies. Therefore, integrating OpenAI Gym with HDDL creates a unified framework for designing and evaluating hierarchical RL approaches, combining RL's adaptive learning with the structured decision-making of hierarchical planning.

%% file: sections/3_related_works.tex
\section{Related Work}
\label{sec:relatedworks}

% [Discuss PDDLGym, a similar tool of converting PDDL domains and problems into Gym environment. Mention that HDDL is more complicated with the hierarchical task networks, so a special implementation is needed to handle hierarchy pruning and selection.]

PDDLGym \cite{silver_pddlgym_2020} constructs Gym environments from PDDL domains and problems, serving as a valuable reference for our work. However, HDDL significantly differs from PDDL, particularly in managing hierarchical task networks or task and method operators. Additionally, PDDLGym operates under a single-action-per-step model, which suits many PDDL domains but lacks the complexity needed for advanced applications, such as multi-agent contexts. In contrast, our framework, HDDLGym, is designed to accommodate multi-agent environments, enabling the study of RL policies in more complex settings.

Similarly, pyRDDLGym \cite{taitler_pyrddlgym_2024} integrates a planning domain language, Relational Dynamic Influence Diagram Language (RDDL) \cite{rddl}, with Gym. RDDL is adept at modeling probabilistic domains with intricate relational structures. However, it does not inherently support multi-level actions. This limitation requires significant adjustments when defining hierarchical problems within PyRDDLGym. Users must creatively structure RDDL descriptions to represent sequences of actions, which can complicate the modeling of hierarchical tasks.

NovelGym is a versatile platform that supports hybrid planning and learning agents in open-world environments \cite{goel2024novelgym}. It effectively combines hierarchical task decomposition with modular environmental interactions to facilitate agent adaptation in unstructured settings. Nevertheless, its hierarchical structure is relatively straightforward, primarily relying on primitive and parameterized actions defined in PDDL. Conversely, HDDLGym offers more advanced hierarchical capabilities through HDDL, granting users greater flexibility and complexity in specifying high-level strategies and problem-solving approaches.

% TODO:
% [Add a paragraph describe what extra implementation that HDDLGym has]

HDDLGym implements several critical extensions to support hierarchical and multi-agent planning. As outlined in the Introduction, its five key features enable users to systematically design and study hierarchical planning in conjunction with RL approaches. In addition, the framework integrates visualization tools and evaluation metrics to facilitate both qualitative and quantitative policy analysis. These enhancements allow HDDLGym to capture the full complexity of hierarchical decision-making in multi-agent environments, enabling capabilities that prior frameworks have not fully supported.

% In conclusion, while prior Gym-based frameworks such as PDDLGym, PyRDDLGym, and NovelGym provide valuable foundations for integrating planning and learning agents, HDDLGym advances the field by offering a robust platform for managing complex, hierarchical, multi-agent environments, thus enabling new lines of research in hierarchical reinforcement learning and multi-agent coordination.

% In conclusion, while prior Gym-based frameworks like PDDLGym, PyRDDLGym, NovelGym, etc. provide valuable foundations for working with planning and learning agents, HDDLGym contributes to the field with its ability to manage complex multi-agent hierarchical environments.

%% file: sections/4_formal_framework.tex
\section{Formal Framework}
\label{sec:formal}
% [include definition from Holler paper and add a set of Agent to it]
Due to various differences in the original formalities and purposes between HDDL and Gym, some modifications in HDDL domain files are required to enable HDDLGym to work smoothly. In this section, we introduce the agent-centric extension of HDDL, modified from the standard HDDL by \citet{holler2020hddl}. The agent-centric extension only includes changes to the HDDL domain. The agent-centric planning domain is defined below:

\begin{definition}
\label{def:domain}
An agent-centric planning domain $\mathcal{D}$ is a tuple 
% $(A, L, T_P, T_C, M)$ 
$\mathcal{D}= \langle t_a, L, T_P, T_C, M \rangle$, where:
\begin{itemize}
    \item $t_a$ is an agent type hierarchy in the domain.
    \item L is the underlying predicate logic.
    \item $T_P$ is a finite set of primitive tasks, also known as actions. Actions can be further classified into agent actions and environment actions.
    \item $T_C$ is a finite set of compound tasks.
    \item $M$ is a finite set of decomposition methods with compound tasks from $T_C$ and task networks over the name $T_P \cup T_C$.
\end{itemize}
\end{definition}

We next discuss the elements in Def.~\ref{def:domain} that are different from the definition of planning domain in Sec.~\ref{sec:background}.

\paragraph{Agent type hierarchy $t_a$}
One major difference compared to the standard HDDL \cite{holler2020hddl} is the addition of $t_a$. $t_a$ is used to specify which types are classified as agent types within the domain.
In an HDDL domain, this classification is done by defining the type ``agent'' within the  \texttt{:types} block. 
% For example, in Transport domain, vehicle is specified as an agent type in the last line of the following types block.
For instance, in the Transport domain, the ``vehicle'' is designated as an agent type, as shown in the line 5 of the types block below. This approach allows the domain to clearly differentiate agent types from other entities, enabling more structured interactions within hierarchical planning tasks.
\begin{quote}
\begin{scriptsize}
\begin{Verbatim}[commandchars=\\\{\},numbers=left]
(:types
        location target locatable - object
        vehicle package - locatable
        capacity-number - object
        \userinput{vehicle - agent})
\end{Verbatim}
\end{scriptsize}
\end{quote}

\paragraph{Primitive Task Set $T_P$}
The primitive task set, \textit{$T_P$}, encompasses all actions defined within the domain, classified as either agent actions or environment actions. Agent actions include one or more agents as parameters, while some actions - initially defined without agent parameters due to the nature of their predicates - must be modified to include agents if these actions are performed on behalf of agents. 
Additionally, in RL, particularly in multi-agent settings, it is essential to ensure that the domain includes a \emph{none} action for each agent, enabling an agent to choose no action for a given step. Therefore, the HDDL domain file should incorporate the following action block to support the \emph{none} action functionality.

\begin{quote}
\begin{scriptsize}
\begin{Verbatim}[numbers=left]
(:action none
    :parameters (?agent - agent)
    :precondition ()
    :effect ())
\end{Verbatim}
\end{scriptsize}
\end{quote}

On the other hand, environment actions exclude agents from their parameters, making them non-agent actions. These actions execute automatically as soon as their preconditions are met immediately after all agents have completed their actions, enabling flexible environment dynamics.

\paragraph{Compound Task Set $T_C$}
The compound task set, \textit{$T_C$}, includes all high-level tasks, aligning with the standard HDDL structure as described by \citet{holler2020hddl}. However, in HDDLGym’s implementation, additional task definition details are required. Specifically, to ensure task completion, HDDLGym checks the current world state against the defined task effects. Thus, task definitions must include explicit effects. In the following example from the Transport domain, the text in bold highlights the revisions made to the original HDDL task definition.

\begin{quote}
\begin{scriptsize}
\begin{Verbatim}[commandchars=\\\{\},numbers=left]
(:task get-to
    :parameters (\userinput{?agent - agent} ?dest - location)
    \userinput{:effect (at ?agent ?destination)})
\end{Verbatim}
\end{scriptsize}
\end{quote}

The remaining components in the tuple, \textit{L} and \textit{M}, are consistent with the standard HDDL formulation as defined by \citet{holler2020hddl}. Note that to facilitate the modification and verification of HDDL domains for compatibility with HDDLGym, the codebase includes an interactive platform featuring autonomous task effect generation, agent parameter augmentation for actions, and related capabilities.
% \footnote{Refer to the IPython notebook tutorial in the code for details.}

%% file: sections/5_hddlgym_framework.tex
\section{HDDLGym Framework}
\label{sec:framework}
This section explains the design and implementation of HDDLGym. It covers (1) details of HDDLEnv as a Gym environment, (2) the definition of the Agent class, (3) observation and action space details, (4) RL policy, (5) planning in multi-agent scenarios, and (6) the HDDLGym architecture.
% First, we start with implementing HDDLEnv, which maps HDDL domain and problem files into a Gym environment. 
% Next, section \ref{subsec:agent} provides details of how an agent is defined and operated in the system.
% Section \ref{subsec:observation} discusses the observation and action spaces of the system, which are critical to define an RL policy in hierarchical planning problems.
% Then, section \ref{subsec:policy} entails how RL policy is designed and implemented in the HDDLGym.
% Next, section \ref{subsec:planner} explains the hierarchical planner of HDDLGym that supports multi-agent settings. Lastly, section \ref{subsec:architecture} demonstrates how all components are combine together in the HDDLGym with the high-level architecture in Figure \ref{fig:architecture}. 

\subsection{Gym and HDDLEnv}
\label{subsec:hddlenv}
% [Gym: describe general components of Gym environment]
In the HDDLGym framework, we introduce HDDLEnv, a Python class that extends the Gym environment to support hierarchical planning with HDDL. 
% Details of the HDDLEnv implementation are available in the \texttt{hddl\_env.py} file.
% Gym environment \cite{brockman_openai_2016} includes methods to initialize the environment, reset the environment, step through the environment. In HDDLGym framework, HDDLEnv is a Python class with all features of a Gym environment. Details of HDDLEnv implementation can be found in file \texttt{hddl\_env.py} of the code.
\paragraph{Initialize and reset functions}
HDDLEnv is initialized using HDDL domain and problem files, together with an optional list of policies for all agents. During initialization, the \texttt{main\_parser} function converts HDDL files into an environment dictionary, setting up the initial state and goals. Agents are then initialized with their associated policies.
% In the \texttt{\_\_init\_\_} function, the \texttt{main\_parser} function is called to convert HDDL files into an environment dictionary. Then, it initializes agents with associated policies.

The reset function optionally accepts a new or updated list of agents' policies and resets the environment to its initial state and goal tasks as specified in the HDDL problem file. It also re-initializes agents with their associated policies. 

\paragraph{Step function}
The step function in HDDLEnv accepts an action dictionary from the agents and returns the new state, reward, `done' flag (indicating win or loss), and debug information, similar to the format of OpenAI Gym's step function. After executing agents' actions, it also checks and applies any valid environment actions. Environment actions are any actions that are not associated with any agent. This design enables the environment to change independently from agents' behaviors. If the current state of the environment satisfies the precondition of an environment action, that action is executed automatically.
\subsection{Agent}
\label{subsec:agent}
% [Mention agent-centric design of HDDLGym]
HDDLGym is designed as an agent-centric system. It inherently focuses on the interactions and actions of agents within the environment. Therefore, defining an Agent class, as in Definition \ref{def:agent} below, is critical in implementing HDDLEnv. 
\begin{definition}
\label{def:agent}
An agent $A$ is defined with a tuple $\langle N, P, B, H, U \rangle$ where: 
\begin{itemize}
    \item $N$ is agent name,
    \item $P$ is a policy,
    \item $B$ is set of agents, representing the agent's belief about other agents' configuration in the environment,
    \item $H$ is a list of tasks, methods, and action, representing the action hierarchy of the agent,
    \item $U$ is a function to update the action hierarchy of agent based on the current state of the world.
\end{itemize}
\end{definition}
% [TODO: Briefly explain each component, using example with Transport or Overcooked]
\paragraph{Initialize an agent}
% [Initializing an agent: key elements: task\_method\_hierarchy, belief\_other\_agents, agent\_policy] 
All agents in the environment are initialized when an HDDLEnv instance is created or reset. Each agent is initialized with a name $N$ and a policy $P$. The agent's name $N$ is derived directly from the HDDL domain and problem files. The policy $P$ refers to an RL strategy that the agent employs to support its hierarchical planning process. 
%This initialization framework enables the agent to operate autonomously, with clearly defined parameters and behaviors, while accounting for both its own actions and the predictions of others within a multi-agent environment.
% This initialization ensures that the agent is properly configured to operate within the hierarchical planning framework.
This initialization configures the agent to operate within the hierarchical planning framework.

\paragraph{Function $U$: updates agent's action hierarchy $H$}
% [How and when to update agent's hierarchy, update hierarchy of belief agents]
An important method in the Agent class is the update hierarchy function $U$. This function checks whether any tasks or actions in the agent's hierarchy $H$ have been completed by comparing their effects with the current state of the world. Once tasks or actions are completed, they are removed from both the agent's hierarchy $H$ and the agent's belief about other agents' action hierarchies ($B$). $U$ is called for each agent after the environment's step function is executed, ensuring that the agents are prepared to plan the next step.

\subsection{Observation and Action Spaces}
\label{subsec:observation}
\paragraph{Observation space}
In general multi-agent problems, each agent can be assumed to have knowledge about the current state of the world, its own hierarchical actions, and other agents' previous actions. Different RL methods have different designs for which information should be included in the inputs and outputs of the RL policies. For example, in the default setup, we set the observation of each agent to include information on (1) current state of the world, (2) goal tasks, (3) the agent's action hierarchy, and (4) other agents' previous primitive actions. Meanwhile, the RL model should return information about the action hierarchy of each agent.

In our current design, we use dynamic grounded predicates to represent the current state of the world. Dynamic grounded predicates represent a subset of all possible grounded predicates within the environment. In HDDL, and PDDL more broadly, predicates can either be static or dynamic. Static predicates define unchanging world conditions (e.g., spatial relationships between locations), while dynamic predicates represent changing world conditions (e.g., agent positions). Dynamic predicates can be added or removed from the world state by actions. 
% To identify all grounded dynamic predicates, we first locate lifted dynamic predicates within an action's add and delete effects, then generate the full set of grounded dynamic predicates from these lifted forms.

% define static world conditions, like the spatial relationships between locations, or dynamic states, such as the positions of agents. The term "dynamic" suggests that these predicates may be added or removed from the world state by actions. To identify all grounded dynamic predicates, we first locate all lifted dynamic predicates within action's add and delete effects, then generate a complete list of grounded dynamic predicates from these lifted forms.

% Goal tasks are specified in the HDDL problem file under the \texttt{:htn} Section. Each agent's action hierarchy is structured as a list, starting with a goal task and ending with a primitive action. Figure 1 illustrates examples of action hierarchies in the two environments, Transport and Overcooked, that are further discussed in Sec. ~\ref{sec:demonstration}.

Our default setup focuses on using dynamic grounded predicates, rather than the full set of grounded predicates, to reduce the observation space. This scalability trade-off is illustrated in Table \ref{table:dimensions}, specifically in Overcooked domain. However, this design choice may limit the generalizability of the RL policy, as it is tailored to a specific set of HDDL problem instances and may not transfer well to problems with different agents, objects, and/or static world conditions. To address this, HDDLGym also allows users to customize the state representations to accommodate diverse needs across domains, for example, using grounded predicates in the Transport domain, where their dimensions remain manageable as the problem size increases (Table \ref{table:dimensions}).

% To represent goal tasks and action hierarchies, a practical method is to one-hot encode grounded operators using a comprehensive list of all possible grounded operators. However, this approach results in a large observation space due to additional operators for tasks and methods in hierarchical problems. Each grounded operator (whether a task, method, or action) can be decomposed into a lifted version paired with relevant objects. This allows agents' goal tasks and action hierarchies to be combined and represented as a one-hot encoded vector based on all possible lifted operators and associated objects. 

\paragraph{Action space}
Unlike PDDL or non-hierarchical planning problems, HDDLGym aims to provide not only primitive actions but also the full action hierarchies that reflect high-level strategies guiding agent behavior. As shown in Table \ref{table:dimensions}, the set of all possible grounded operators can grow prohibitively large in complex domains, while lifted operators offer a more compact alternative. A middle-ground approach uses lifted operators with associated objects to retain contextual information. Our default setup uses the middle-ground approach. The action hierarchy of each agent is one-hot encoded over all lifted operators and objects, reducing observation and action space sizes by omitting subtask orders and specific object-operator links. HDDLGym also supports flexible state and action space designs, including multiple RL models for different operator types and customizable encoding schemes for diverse domains and experiments.

\begin{table}[t]
\centering
\resizebox{1\columnwidth}{!}{
\begin{tabular}{l c c c c c }
\toprule
    \textbf{Domain} & \multicolumn{3}{ c }{\textbf{Transport}} & \multicolumn{2}{ c }{\textbf{Overcooked}}\\
    \cmidrule(lr){1-1}\cmidrule(lr){2-4}\cmidrule(lr){5-6}
    \textbf{\# of Agents} & \textbf{1} & \textbf{2} & \textbf{3} & \textbf{2} & \textbf{3}\\
    \midrule
    G. predicates & 35 & 79 & 269 & 937,158 & 1,186,066\\
    G. dynamic pred. & 13 & 38 & 132 & 90 & 101\\
    \midrule
    G. operators & 184 & 1570 & 11,979 & 200,597 & 300,860\\
    G. actions & 58 & 610 & 4,093 & 175,791 & 263,668\\
    L. operators & 14 & 14 & 14 & 18 & 18\\
    L. actions & 4 & 4 & 4  & 5 & 5 \\
    Objects & 8 & 13 & 23  & 18 & 19\\
    \bottomrule

\end{tabular}}
\caption{\textbf{Dimensions of Lifted and Grounded Representations in Transport and Overcooked Problems.} \textit{G.} denotes Grounded, and \textit{L.} denotes Lifted. The large number of grounded predicates in Overcooked highlights the need to use dynamic grounded predicates for state representation. Likewise, grounded operators are impractical for defining the action space in RL model training due to their scale.}
\label{table:dimensions}
\end{table}

% [Compare to PDDLGym: free param technique]
% Another approach to reduce the size of the action space is explored in PDDLGym~\cite{silver_pddlgym_2020}, where they introduce 
% a distinction between \textit{free} and \textit{non-free} parameters. Free parameters convey the essential information of an action, while non-free parameters are included due to their presence in precondition or effect expressions. Consequently, PDDLGym's action space consists of combinations of lifted operators with their free parameters. Although this approach works well in PDDLGym, it is challenging to implement within the HDDLGym framework, as identifying free parameters for tasks and methods is not trivial.

% The reducing the size of the action space was also explored in the PDDLGym project \cite{silver_pddlgym_2020}. This framework introduced the distinction between free and non-free parameters. Free parameters convey the essential information of an action, while non-free parameters are included due to their presence in precondition/effect expressions. Consequently, PDDLGym's action space comprises combinations of lifted operators with their free parameters. Although this approach is effective in PDDLGym, it is challenging to apply within the HDDLGym framework because identifying free parameters for tasks and methods is not straightforward.

\subsection{RL Policy}
\label{subsec:policy}
% [Describe RL policy: input and output, how it support hierarchical planner]
The RL policy plays a crucial role in the HDDLGym framework by supporting the search for an optimal hierarchical plan for each agent. In the default setup, the policy takes the observation as input, which includes information about dynamic grounded predicates, goal tasks, and previous action hierarchies. Its output is the probabilities of lifted operators and objects, which are then used to compute the probabilities of grounded operators. These probabilities guide the search for action hierarchies within the HDDLGym planner, as discussed in Sec.~\ref{subsec:planner}.
% the following section.

In this work, we implemented Proximal Policy Optimization (PPO) \cite{ppo} for discrete domains to effectively explore the application of RL in hierarchical planning problems. 
% $TODO: Mention more RL designs$
HDDLGym is designed to enable users to flexibly integrate their preferred RL models, including multi-layer perceptrons (MLPs), recurrent neural networks (RNNs), and others. Comprehensive guidance and detailed examples are provided in the tutorial Python notebook included in the codebase.
% Details about the RL policy (including input-output spaces, training methods, and evaluation processe) are organized in the \texttt{learning\_methods.py} file. By modifying this Python file, users can easily experiment with and implement alternative RL frameworks, supporting further research in multi-agent hierarchical planning.
% Other RL methods can also be integrated into the HDDLGym framework, provided they implement similar functions such as \texttt{select\_action}, \texttt{train}, \texttt{put\_data}, \texttt{save}, and \texttt{load}, as described in \texttt{learning\_methods.py} of the code.

\subsection{Planning for Multi-agent Scenarios}
\label{subsec:planner}
HDDLGym is designed to work in multi-agent settings; therefore, the planner also considers collaboration between agents. The HDDLGym planner is designed in a centralized format. In decentralized planning, each agent runs the centralized planner using its own information and beliefs about the other agents.
% Details of HDDLGym planner implementation can be found in \texttt{central\_planner.py} and \texttt{central\_planner\_utils.py}.

% The HDDLGym Planner algorithm, Algorithm 1, describes how HDDLGym find the next action hierarchies for agents.
Algorithm 1 
% defines
outlines the approach of
the HDDLGym Planner, where agents determine their action hierarchies by iteratively updating through valid operator combinations. 
Particularly, HDDLGym Planner's inputs are list $\mathcal{A}$ of all agents with uncompleted hierarchies, policy \textit{P}, and deterministic flag \textit{d}. The HDDLGym planner is a centralized planner. In case of decentralized planning, the list $\mathcal{A}$ include a real agent and that agent's belief about other agents. The deterministic flag \textit{d} determines whether the selection process should follow a deterministic or probabilistic approach when choosing operators to form agents' action hierarchies. The policy \textit{P} guides the search for a suitable hierarchy based on the flag $d$. 
% In the codebase, \textit{P} may comprise multiple sub-policies applied at different planning layers. For simplicity, \textit{P} can be treated as a single RL policy here. 
The HDDLGym Planner outputs an updated list of agent instances, \textit{$\mathcal{A}_{updated}$}, with each agent's action hierarchy terminating in a grounded primitive action.

The planner begins by initializing an empty list, \textit{Done}, to keep track of agents whose hierarchies end with an action (line 1). The while loop from lines 2 to 28 continues until all agents have completely updated their hierarchies. Within this loop, an empty list, \textit{$O_\mathcal{A}$}, is initialized to store the valid operators of all agents in \textit{$\mathcal{A}$} (line 3). Next, the for-loop from lines 4 to 17 iterates to find all valid operators \textit{$O_a$} for each agent \textit{a}, for \textit{a} $\in \mathcal{A}$. To do this, the algorithm first checks if \textit{a} is in \textit{Done}, meaning its hierarchy is complete (line 5). If so, then \textit{$O_a$} is set as a list containing the agent \textit{a}'s primitive action (line 6). Otherwise, the while loop from lines 8 to 14 runs until it finds a non-empty \textit{$O_a$}. In this while loop, the list of valid operators for \textit{a} is validated in line 9; if no valid operators are found (line 10), the last operator in \textit{a}'s hierarchy is removed, and the loop is rerun. However, if \textit{a}'s hierarchy is already empty, indicating that no valid operator can be found for \textit{a}, the \texttt{none} action is added to \textit{$O_a$} (line 12).

The operator list \textit{$O_a$} for each agent is then added to \textit{$O_\mathcal{A}$}, the list of operators for all agents (line 16). This list, \textit{$O_\mathcal{A}$}, is subsequently used to generate all combinations of joint operators, \textit{C} (line 18). Line 19 details the pruning of invalid combinations in \textit{C}. A combination is invalid if it violates either of two conditions: first, no agent should perform multiple different actions; and second, no action in the combination should have effects that conflict with the preconditions of other actions. After this pruning, \textit{C} contains only valid operator combinations.

Lines 20 to 25 describe how the policy \textit{P} is applied to select a combination \textit{c} from the list of valid combinations, \textit{C}. The probability list, \textit{$P_O$}, corresponding to \textit{C} is generated using policy \textit{P}. Depending on the deterministic flag \textit{d}, the chosen combination \textit{c} is selected in either deterministically (line 22) or probabilistically (line 24).

With the combination of operators determined, the next step is to use it to update each agent’s hierarchy (line 26). The list \textit{Done} is then updated if any agents have completed hierarchies (line 27). This process is repeated until all agents in $\mathcal{A}$ have completed their hierarchies. At this point, the HDDLGym planner returns the list of fully updated agents, $\mathcal{A}_{updated}$, as shown on line 29.

\begin{algorithm}[tb]
\caption{HDDLGym Planner}
\label{alg:planner}
% \textbf{Input}: $real\_agent\_index$, $all\_agents$, $all\_policies$, $is_deterministic$\\
\textbf{Input}: list of agents $\mathcal{A}$, deterministic flag \textit{d}, policy \textit{P} \\
\textbf{Output}: updated list of agents $\mathcal{A}$
\begin{algorithmic}[1] %[1] enables line numbers
\STATE Initialize an empty list \textit{Done} to keep track of agents whose hierarchies reached action.
\WHILE{not all agents in \textit{Done}}
\STATE Initialize an empty list \textit{$O_\mathcal{A}$} for valid operators of $\mathcal{A}$
\FOR{agent \textit{a} in $\mathcal{A}$}
\IF{\textit{a} in \textit{Done}}
\STATE \textit{$O_a$} $\leftarrow$ [action of agent \textit{a}]
\ELSE
\WHILE {\textit{$O_a$} not empty}
\STATE \textit{$O_a$} $\leftarrow$ a list of valid operators for \textit{a}
\IF{\textit{$O_a$} is empty}
\STATE Remove the last operator of agent \textit{a} hierarchy from its hierarchy
\STATE If no more operator from \textit{a}'s hierarchy to remove, add \texttt{none} action to \textit{$O_a$} 
\ENDIF
\ENDWHILE
\ENDIF
\STATE Add \textit{$O_a$} to \textit{$O_\mathcal{A}$}
\ENDFOR
\STATE \textit{C} $\leftarrow$ Combinations of joint operators from \textit{$O_\mathcal{A}$}
\STATE Remove any invalid combinations from \textit{C}
\STATE $P_O \leftarrow $ Probability of combinations in \textit{C} with \textit{P}
\IF {\textit{d} is True}
\STATE $c \leftarrow \emph{argmax}_{c \in C} P_O $
\ELSE
\STATE $c \leftarrow$ Randomly from $C$ with weights be $P_O$
\ENDIF
\STATE Update hierarchies of all agent $\mathcal{A}$ with operators in $c$
\STATE Check each agent's hierarchy and update \textit{Done} if any hierarchy ends with action
\ENDWHILE
\STATE \textbf{return} $\mathcal{A}_{updated}$
\end{algorithmic}
\end{algorithm}

\subsection{HDDLGym Architecture}
\label{subsec:architecture}
\begin{figure}[t]
\centering
\includegraphics[width=0.99\columnwidth]{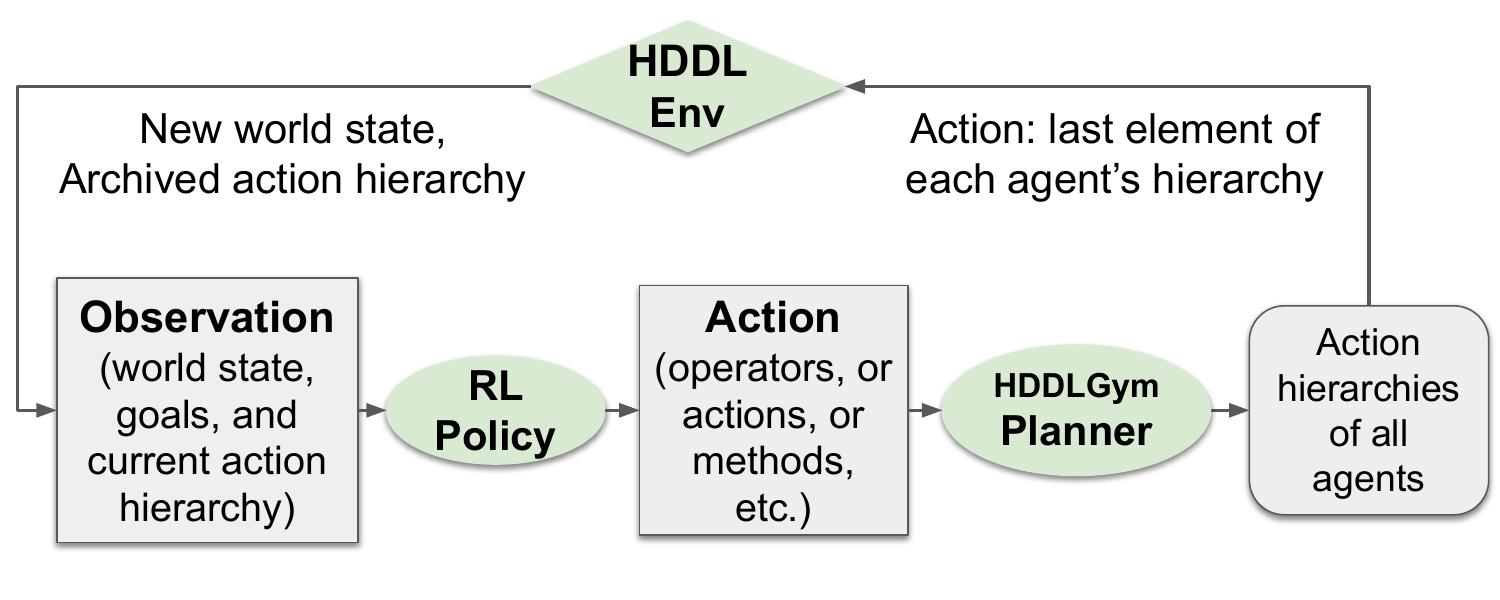} % Reduce the figure size so that it is slightly narrower than the column. Don't use precise values for figure width.This setup will avoid overfull boxes.
\caption{\textbf{HDDLGym high-level architecture.} Outputs of RL policy help HDDLGym Planner update the action hierarchy of each agent. Then, primitive actions are extracted from the hierarchies, and applied to the environment.}
\label{fig:architecture}
\end{figure}

The high-level architecture of HDDLGym is demonstrated in Figure \ref{fig:architecture}. As discussed in Section \ref{subsec:observation}, 
% the input of the RL policy is a one-hot encoded vector that includes (1) dynamic grounded predicates, (2) lifted operators, and (3) objects representing the goal tasks and previous action hierarchies of all agents. The output of RL policy is the probabilities of lifted operators and objects. 
% From output list of probabilities, we calculate the probabilities of the grounded operators by averaging the log-probabilities of the lifted operators and the objects involved in the grounded operators. 
The RL policy, described in Section \ref{subsec:policy}, takes an observation as input and outputs a probability distribution over action representations (see Section \ref{subsec:observation}). These probabilities guide the HDDLGym planner in selecting the most appropriate action hierarchy for each agent, as outlined in Algorithm \ref{alg:planner} and Section \ref{subsec:planner}. Primitive actions are then extracted from the updated hierarchies and executed in the environment, resulting in a new world state via the step function (Section \ref{subsec:hddlenv}). Completed tasks or actions are archived and removed from the each agent's action hierarchy $H$ with method $U$ (Section \ref{subsec:agent}) before proceeding to the next cycle. This integrated process supports dynamic and adaptive agent behavior based on both learned policies and hierarchical planning.

%% file: sections/6_applications.tex
\section{Applications}
% \section{Domain Configurations}
\label{sec:applications}

% In this section, we discuss the two classes of domains that are supported in HDDLGym: the IPC-HTN and OpenAI Gym-based domains. 
Table \ref{table:domains} lists the domains, primarily from IPC-HTN \cite{IPC_HTN_tracks} and custom designs, that have been studied with HDDLGym and are included in the codebase\footnote{This list will be updated as the tool evolves. For the most recent version of this work, please refer to the latest arXiv version or visit: https://ngocla.github.io/files/HDDLGym.pdf}. We highlight two representative examples: Transport from IPC-HTN and Overcooked, a popular multi-agent problem in OpenAI Gym.
% \footnote{More domains in IPC-HTN are included in the codebase of the system. More domains from OpenAI Gym will be released with the final version of the paper.}
\begin{table}[t]
\centering
\resizebox{1\columnwidth}{!}{
\begin{tabular}{l c c c c}
\toprule
    \textbf{Domain} & \textbf{Source} & \textbf{Agent-centric} & \textbf{Collab} \\
    \midrule
    Transport & IPC-HTN & Yes & No \\
    Transport Collab. & Modified IPC-HTN & Yes & Yes \\
    Overcooked & Ours & Yes & Yes \\
    Rover & IPC-HTN & Yes & No \\
    Satellite & IPC-HTN & Yes & No \\
    Depots & IPC-HTN & Hidden & No \\
    Minecraft-Player & IPC-HTN & Hidden & No \\
    Barman-BDI & IPC-HTN & Hidden & Yes \\
    Search and Rescue & Ours & Yes & Yes \\
    ZenoTravel & IPC-HTN & Yes & No \\
    Taxi & Ours & Yes & No \\
    Factories-simple & IPC-HTN & No & No \\
    \bottomrule
    
\end{tabular}}
\caption{\textbf{List of Domains Included in the HDDLGym Codebase (As of June 2025)}. \textit{Agent-centric} domains are those in which the agent can be explicitly identified. \textit{Hidden} indicates that the domain is agent-centric, but the agent is not specified as an explicit parameter.}
\label{table:domains}
\end{table}
% \footnotetext{This list will be updated as the tool evolves. For the latest version of the work, please visit: https://ngocla.github.io/files/HDDLGym.pdf}

\subsection{IPC-HTN Domains}
As previously discussed, Gym defines interactions between agents and the environment. Therefore, not all HDDL domains from IPC-HTN  \cite{IPC_HTN_tracks} are directly compatible with HDDLGym. Since agent specification within a domain is necessary, this requirement may not be feasible or appropriate for every IPC-HTN domain. HDDLGym is particularly well-suited to domains with agent-centric systems, such as Transport (where the vehicle serves as the agent), Rover (with the rover as the agent), and Satellite (with the satellite as the agent). To better illustrate the applications of these agent-centric environments, we provide a detailed discussion on Transport domain as follows.
% To better illustrate how existing HDDL domains and problems from IPC-HTN can be utilized with HDDLGym, we provide a detail demonstration of Transport domain and problems.
% Some applicable HDDLGym domains from the 2023 IPC-HTN domain resource \cite{IPC_HTN_tracks} include Depots, Factories-simple, Hiking, Logistics-Learned-ECAI-16, Rover, Satellite, Snake, and Transport.

% Before running HDDLGym, ones must modify the domain files as follows:
% \begin{itemize}
%     \item Specify agent type: explicitly define 'agent' as a type or supertype, and ensure there are objects in the problem file with the agent type.
%     \item Verify agent actions: For all agent actions, agent type at least appear once in parameters.
%     \item Add "none" action for agent: as described in the "None action for agent" part of the previous section.
%     \item Add effect to task: as described in the "Add effect to task" part of the previous section.
    
% \end{itemize}

% Here are steps to run the existing HDDL domains and problems with HDDLGym:
% \begin{itemize}
%     \item Set up the HDDLGym as described in the Installation section
%     \item Modify the existing HDDL domain files to meet the format criteria mentioned in the HDDL Domain Requirements section
%     \item Add any collaborative operators to study interesting features in multi-agent contexts
%     \item Train policy for particular domain and problem files.
% \end{itemize}

\subsubsection {Transport domain}
% To better illustrate how existing HDDL domains and problems from IPC-HTN can be utilized with HDDLGym, we provide a detail demonstration of Transport domain and problems.
%[Describe Transport domain]
The goal of a Transport problem is to deliver one or more packages from their original locations to designated locations.

% To run the Transport domain with HDDLGym, several modifications as described in section \ref{sec:formal} should be followed first. In the Transport domain, the agent is designated as `vehicle'. All actions in the original Transport domain file originally contain vehicle in their parameters.  Particularly, the block \texttt{:type} should include the line ``\texttt{vehicle - agent}'' to specify agent is a super type of the vehicle type. Additionally, the none action for the agent is added as described in Section \ref{sec:formal}.
% Another modification is adding effects to all tasks. An example is the bold text in the following task definition.
% \begin{quote}
% \begin{scriptsize}
% \begin{Verbatim}[commandchars=\\\{\}, numbers=left]
% (:task deliver 
%     :parameters (?p - package ?l - location)
%     \userinput{:effect (at ?p ?l)}
% )
% \end{Verbatim}
% \end{scriptsize}
% \end{quote}

% From here, the transport domain and problem files are ready to use HDDLGym to find the hierarchical plan. Action hierarchy results are illustrated in Figure \ref{fig:transport}. In the scenario, the truck completes the ``delivery package'' goal task after four actions. In each step, the truck's action hierarchy start with the goal task, and end with an action. Its hierarchy is updated after each step to follow a sequence of tasks in ``method-deliver'' to accomplish the ``delivery package'' goal task.
% At this point, the Transport domain and problem files are ready for use in HDDLGym to find a hierarchical plan. 
The resulting action hierarchy of an 1-agent Transport problem is illustrated in Figure \ref{fig:transport}. In this scenario, the truck completes the \texttt{delivery package} goal task after four actions. At each step, the truck’s action hierarchy begins with the goal task and concludes with a primitive action. The action hierarchy updates after each step, following a sequence of subtasks in \texttt{method-deliver} to accomplish the \texttt{delivery package} goal.

% [collab transport]
To evaluate the capability of handling collaborative interactions in Transport domain, we embed the collaborative task, method, and action to the Transport domain. Specifically, task \texttt{transfer}, method \texttt{m-deliver-collab}, and action \texttt{transfer-package} are added in the domain to enable the packages to be transferred from one vehicle to another when the vehicles are at adjacent locations.
Details of these collaborative operators can be found in the codebase.
% Additioanlly, illustrative figures of the collaborative Transport domain are provided in the Appendix.
Following this template, users can explore more interesting interactions and modifying Transport domain to study heterogeneous multi-agent problems. 

% [Conclusion for Transport domain, and IPC-HTN in general]
Above is an example of how to modify an existing IPC-HTN domain to study with HDDLGym and explore more interesting features for multi-agent hierarchical planning. A similar process can be applied to other domains such as Rover, Satellite, and Barman-BDI, to plan with HDDLGym in single or multi-agent contexts.
% In our codebase, we include the modified HDDL domain files of Transport, Rover, Satellite, and Barman-BDI to run with HDDLGym. Other domains from IPC-HTN can also be modified as instructed to use with HDDLGym.  

% \subsection{Custom HDDL domains and problems}
\subsection{OpenAI Gym-based Domains}
\label{subsec:gymdomain}
Writing HDDL domains and problems for an environment is not trivial, especially domains with complicated interaction rules. While there are many ways to do so, we suggest starting with the goal task, then designing methods to achieve the goal task, then coming up with other intermediate tasks and methods for them, and gradually working to the primitive actions. Here is an example of how HDDLgym is applied in Overcooked environment \cite{carroll2019utility}.

\subsubsection{Overcooked}

% Overcooked \cite{carroll2019utility} is a popular Gym-based environment for studying reinforcement learning (RL), modeled after the cooperative and fast-paced mechanics of the original game. In Overcooked, players work together to complete cooking tasks under time constraints. In this scenario, two or more chefs must collaborate to prepare an onion soup. To do so, they need to place an onion in a pot, interact with the pot to start cooking, pour the cooked soup into a bowl, and deliver the bowl to the serving station (see Figure \ref{fig:overcooked}).

Overcooked \cite{carroll2019utility} is a popular Gym-based environment for studying RL, modeled after the cooperative, fast-paced mechanics of the original game. Players, acting as chefs (agents), collaborate to prepare an onion soup by placing onions in a pot, cooking, pouring the soup into a bowl, and serving it (see Figure \ref{fig:overcooked}).

\begin{figure}
    \centering
    \includegraphics[width=.9\columnwidth]{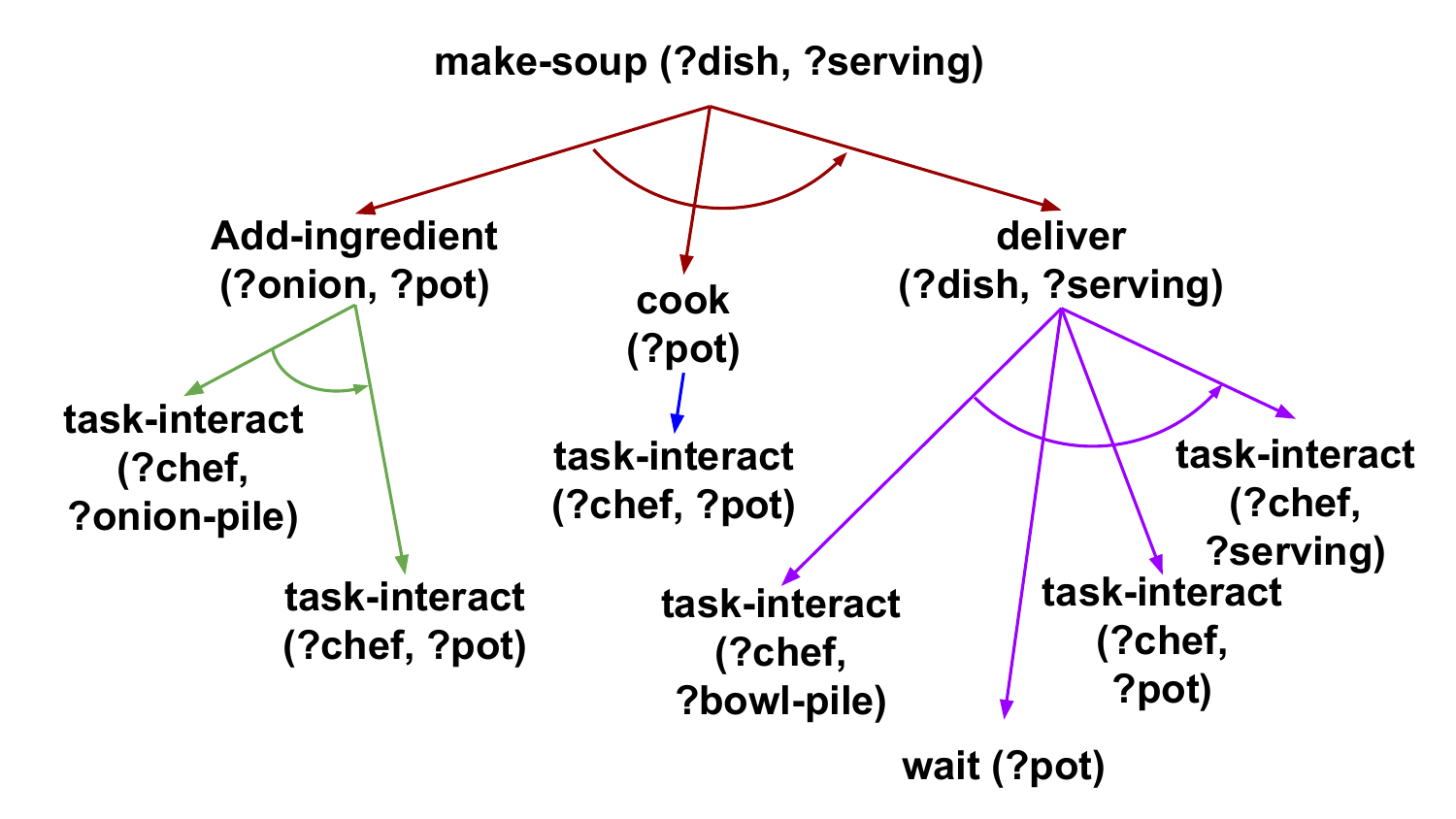} % Adjust width as needed
    \caption{HTNs of the Overcooked domain.}
    \label{fig:htn-overcooked}
\end{figure}

In typical Overcooked scenarios, each agent can perform six primitive actions: moving in a 2D gridworld (up, down, left, right), interacting with objects, or doing nothing. 
% Planning in Overcooked involves both task and motion planning. Motion planning, which operates at a lower level, focuses on spatial relations to determine primitive actions, while task planning identifies the next goal for an agent to approach and interact with. 
Although the whole Overcooked scenarios could be fully defined using HDDL, we found it more efficient to utilize HDDLGym for high-level planning and then apply A$^*$ \cite{duchovn2014path} for motion planning to find the primitive actions as listed above. 
The core HTN for Overcooked domain is entailed in Figure \ref{fig:htn-overcooked}. In the HDDL domain, we define following tasks: make-soup, add-ingredient, cook, deliver, wait, and task-interact. Each of them has one or more method to complete the tasks. Figure \ref{fig:htn-overcooked} only lists several key HTNs of the domain, though all HDDL domain and problem files of Overcooked environment can be found from the codebase. Additionally, Figure \ref{fig:overcooked} shows an example of a hierarchy of an agent and its belief about the other agent's hierarchy.

The following videos help visualize the result of combining HDDLgym in task planning and use A$^*$ for motion planning in various Overcooked layouts:

\begin{links}
    \link{Bottleneck}{https://tinyurl.com/hddlBottleNeckRoom}
    \link{Coord. ring}{https://tinyurl.com/hddlCoordinationRing}
    \link{Left isle}{https://tinyurl.com/hddlLeftIsle}
    \link{Counter circuit}{https://tinyurl.com/hddlCounterCircuit}
    \link{Cramped room}{https://tinyurl.com/hddlCrampedRoom}
\end{links}

% https://tinyurl.com/hddlCoordinationRing
% https://tinyurl.com/hddlLeftIsle
% https://tinyurl.com/hddlCounterCircuit
% https://tinyurl.com/hddlCrampedRoom

\subsection{Evaluation Metrics}
\label{subsec:metrics}
\subsubsection{Complexity and Difficulty}
HDDLGym provides a range of metrics to evaluate RL models within hierarchical planning contexts, with particular focus on plan complexity and planning difficulty. Plan complexity can be assessed by examining the dimensionality of the elements defining a problem, as summarized in Table \ref{table:dimensions}. Complementing this, planning difficulty can be estimated prior to training by measuring planning time, number of planning steps, and success rate using random exploration. Table \ref{table:difficulty} presents such evaluations across the Transport, Overcooked, Rover, and Satellite domains as the number of agents and problem size increase.

% Notably, several domains exhibit a 0\% success rate when the number of agents reaches three or more, highlighting the scalability challenges that arise during training. A low probability of generating successful episodes through exploration increases the risk of training on low-quality data, which may degrade the model’s overall performance.

% These metrics also enable comparison between different hierarchical task networks (HTNs). For instance, in the Transport domain, introducing a collaborative HTN in simple scenarios, such as with one or two homogeneous agents, may add unnecessary complexity and increase planning time. In contrast, more complex settings with heterogeneous agents benefit from collaborative strategies, as omitting them can result in longer plans (e.g., 25 vs. 18 steps) and decreased planning efficiency.

Several domains show a 0\% success rate with three or more agents, highlighting scalability issues during training. Low success probabilities from exploration increase the risk of low-quality data, which can degrade model performance.

These metrics also enable comparisons between different methods, also known as hierarchical task networks (HTNs). For example, in the Transport domain, adding a collaborative method in simple scenarios with one or two homogeneous agents can unnecessarily increase planning time. However, in more complex settings with heterogeneous agents, collaborative strategies improve efficiency, reducing plan lengths (e.g., 25 vs. 18 steps).

In summary, evaluating the complexity and difficulty of the plan in HDDLGym prior to training provides valuable information on the scalability of the domain. This helps users better tailor HTN structures and adjust training strategies, such as reward shaping or extending exploration horizons, to improve convergence and model performance.

\begin{table}[!]
\centering
\resizebox{1\columnwidth}{!}{
\begin{tabular}{l c c c c c}
\toprule
    \textbf{Domain} & \textbf{\# agents} & \textbf{Plan time (sec)} & \textbf{Avg steps.} & \textbf{Success rate} \\
    \midrule
    Transport & 1 & 0.059 & 21 & 100\% \\
    Transport Collab & 1 & 0.077 & 22 & 100\%\\
    Transport & 2 & 0.374 & 41 & 38\% \\
    Transport Collab. & 2 & 1.317 & 35 & 46\% \\
    Transport Collab. & hetero. 2 & 0.795 & 18 & 1\% \\
    Transport & hetero. 2 & 0.367 & 26 & 2\% \\
    Transport & 3 & NA & NA & 0\% \\
    Transport Collab. & 3 & NA & NA & 0\% \\
    \midrule
    Overcooked & 2 & 102.267 & 26& 80\% \\
    Overcooked & 3 & 194.794 & 28 & 40\% \\
    \midrule
    Rover & 1 & 21.360 & 39 & 59\% \\
    Rover & 2 & 87.305 & 38 & 11\% \\
    Rover & 3 & 120.732 & 49 & 2\% \\
    Rover & 4 & NA & NA & 0\% \\
    \midrule
    Satellite & 1 & 0.036 & 12 & 100\% \\
    Satellite & 2 & 1.600 & 21 & 78\% \\
    Satellite & 3 &  6.915 & 25 & 44\% \\
    Satellite & 4 & NA & NA & 0\%\\
    \bottomrule
    
\end{tabular}}
\caption{\textbf{Plan Difficulty of Transport, Overcooked, Rover, and Satellite Problems.} \textit{Hetero.} indicates heterogeneous agents, meaning agents have different capabilities. \textit{NA} means no successful run occur within the maximum episode steps of 100.}
\label{table:difficulty}
\end{table}

\subsubsection{Training}
\begin{figure*}
    \centering
    \includegraphics[width=2\columnwidth]{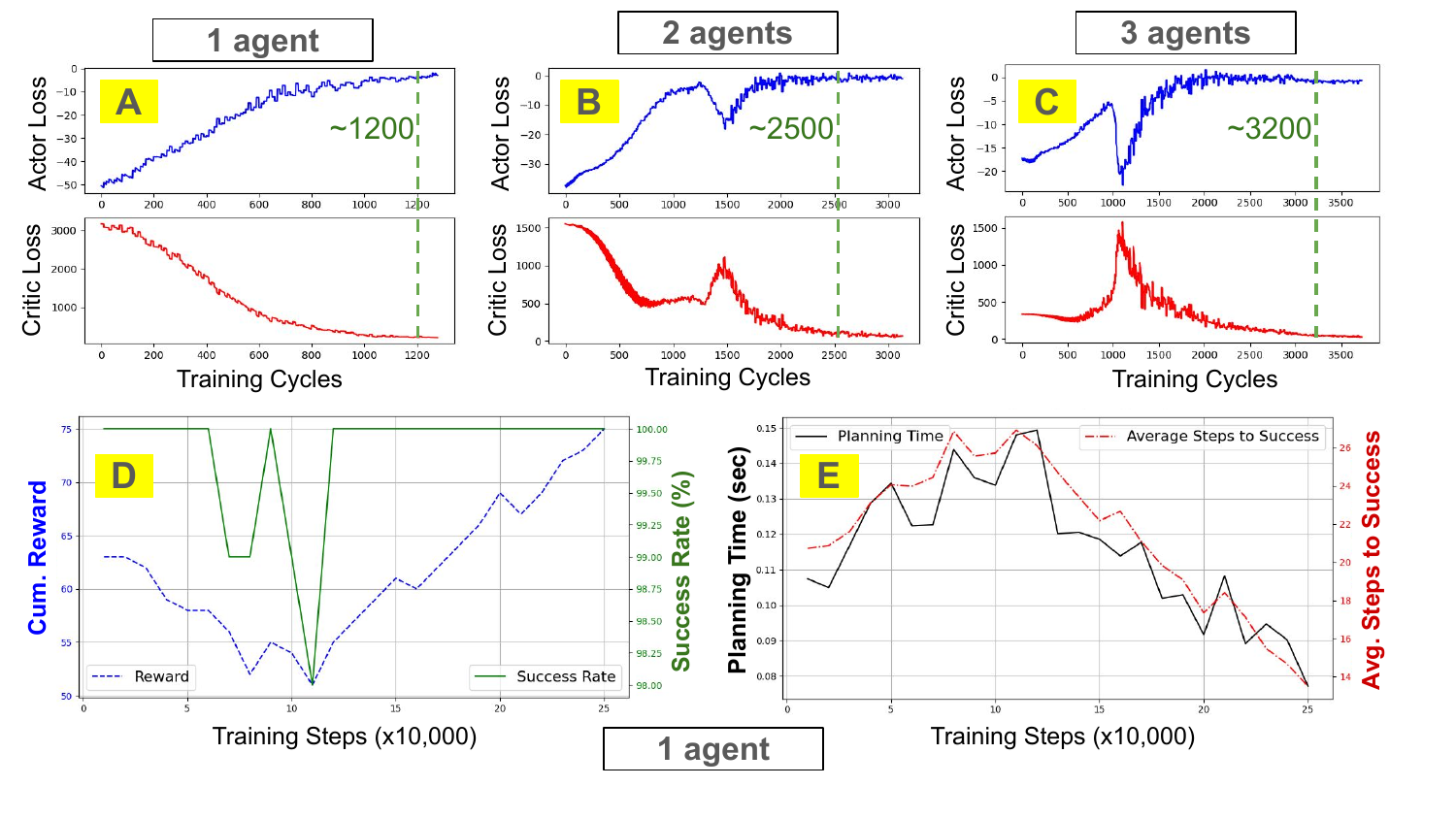} % Adjust width as needed
    \caption{\textbf{Training Dynamics Analysis.} Figures A, B, and C show the training losses for the Transport domain with 1, 2, and 3 agents, highlighting longer convergence times as the number of agents increases. Figures D and E display the PPO policy's training progression for the 1-agent Transport problem, including cumulative reward, success rate, planning time, and steps.}
    \label{fig:training}
\end{figure*}

Several metrics are implemented to evaluate the RL training process. The first is the loss value, which monitors the convergence rate and is illustrated in plots A–C of Figure \ref{fig:training} for the Transport 1-agent, 2-agent, and 3-agent problems. Loss plots are generated and saved at specified intervals during training to help users track the learning progress. Secondly, the policy is periodically evaluated using quantitative metrics such as cumulative discounted reward, success rate, planning time, and planning steps to assess whether the RL policy is converging toward an optimal solution. These evaluation results are recorded in two graphs, as shown for the Transport 1-agent problem in parts D and E of Figure \ref{fig:training}.

% Figure \ref{fig:training} is an example of Transport domain training processes for 1-agent, 2-agent, and 3-agent problems.

\subsubsection{Deployment}
After training, the next step is to deploy and evaluate the RL models. HDDLGym supports a variety of standard quantitative evaluation metrics, including the execution time of the RL-assisted planner, the number of planning steps needed to achieve the specified goals, and the success rate, similar to those reported in Table \ref{table:difficulty}. For qualitative assessment, HDDLGym provides a visualization tool with detailed usage instructions in the codebase, allowing users to examine the action hierarchy of each agent step by step. This visualization can be integrated with the domain renderer, for example, in Overcooked demonstration videos discussed in Section \ref{subsec:gymdomain}, to facilitate a more comprehensive evaluation of agent performance during deployment. Together, these quantitative and qualitative tools support in-depth analysis, helping users refine their models and compare performance across different training settings and domain configurations.

% \begin{figure}
%     \centering
%     \includegraphics[width=\columnwidth]{images/hddl_planner.png} % Adjust width as needed
%     \caption{\textbf{Visualization of action hierarchies at a single step in the Overcooked 2-agent collaborative scenario.} Each agent maintains its own action hierarchy and a belief about the other agent’s hierarchy, reflecting the decentralized planning scheme.}
%     \label{fig:visualization}
% \end{figure}

% \begin{figure}
%     \centering
%     \includegraphics[width=\columnwidth]{images/Overcooked_demo_with_visualization.pdf} % Adjust width as needed
%     \caption{Hierarchy outputs of the Overcooked 2-agent collaborative scenario in the demo.}
%     \label{fig:demo}
% \end{figure}

% These metrics are also useful for comparing the effectiveness of different hierarchical task networks (HTNs), including models trained with and without collaborative structures. For example, in a simple Transport problem where collaboration is unnecessary, incorporating a collaborative HTN may increase planning overhead by introducing additional complexity. Conversely, in more complex Transport problems, the absence of a collaborative strategy may lead to unsolvable scenarios.

% In summary, these evaluation tools offer a comprehensive framework for assessing the scalability, learning efficiency, and practical effectiveness of RL models within hierarchical planning environments.

%% file: sections/7_discussion_future_works_conclusions.tex
\section{Discussion and Future Work}
\label{sec:discussion}

We introduced HDDLGym, which transforms HDDL-defined hierarchical problems into Gym environments, enabling the use of RL policies in hierarchical planning systems. By prioritizing scalability in observation and action spaces, HDDLGym makes trade-offs that enhance complexity handling at the cost of slight accuracy loss in RL models. This flexibility is crucial for tackling intricate tasks. Additionally, HDDLGym supports multi-agent environments, enriching the framework for studying collaborative dynamics in hierarchical planning and offering engaging RL research scenarios.

HDDLGym currently operates under certain limitations that we aim to address in future developments. 
First, it can only handle discrete state and action spaces, which restricts its application to scenarios that require continuous or hybrid spaces. 
% Additionally, HDDLGym’s multi-agent structure is symmetric, meaning all agents have equal roles and no agent has priority over another in task selection. This is a simplification that does not always align with real-world collaborative multi-agent systems, where some agents may have dominant roles or specific priorities. 
Furthermore, HDDLGym assumes a deterministic transition function, meaning that action effects are predictable and do not account for probabilistic outcomes. This limits its applicability to environments where uncertainty and stochastic outcomes are common. 
% Lastly, similar to standard RL setups, the RL policy trained for HDDLGym problems is specific to a particular problem file within a domain and may not generalize effectively to other problem files. Changes such as a varying number of agents, different objects, or altered initial state conditions require retraining the policy, which hinders scalability and adaptability across diverse scenarios within the same domain. Addressing these limitations will be essential to broaden HDDLGym’s usability in complex, real-world settings.
Lastly, the use of one-hot encoding for observations as input to an RL policy restricts its applicability to problems involving similar objects. In many cases, when only dynamic predicates are used for observations, the RL policy is confined to a fixed static condition. Changes like varying agent numbers, adding/removing objects, or altering static conditions require a different RL policy, limiting scalability and adaptability across scenarios within the same domain. Overcoming these challenges will be crucial for expanding HDDLGym’s applicability to complex, real-world settings.

% \subsection{Future Work}
% While converting existing HDDL domains to use with HDDLGym is straightforward, translating native Gym environment into HDDL domain and problem files is not trivial. An on-going work is to trying to converting more agent-centric environments, similar to Overcooked, into HDDL format to exploit the benefits of using HDDLGym.
% While converting existing HDDL domains for use with HDDLGym is relatively straightforward, translating native Gym environments into HDDL domain and problem files is significantly more complex. Current efforts focus on converting more agent-centric environments, such as Overcooked, to the HDDL format to leverage HDDLGym’s advantages. This ongoing work aims to expand the compatibility of agent-based Gym environments with HDDLGym, enabling more complex multi-agent hierarchical planning applications. 
% In the future, HDDL domains can also be learned autonomously using recent advances in the field of learning HDDL domains from observations~\cite{grand2022accurate}.
In the future, HDDL domains can be learned autonomously through advances in: (i) offline learning of HDDL domains from observations~\cite{grand2022accurate}, (ii) offline learning of HTNs from observations~\cite{li2014learnhtn,ZHUO2014134,liteaching}, (iii) online learning of PDDL-like domains~\cite{Ng2019Incremental,Lamanna2021Online,verma2021asking,verma2022discovering}, and (iv) online learning of HDDL domains from human input with Large Language Models~\cite{fine2024leveraging,favier2025leveraging}.

As discussed, HDDLGym has limitations that could be addressed to better support complex multi-agent hierarchical problems. An improvement is enabling HDDLGym to handle multiple pairs of HDDL domain and problem files for different agents within a single Gym environment. Inspired by how multi-agent features are added to PDDL and HTNs through MA-PDDL \cite{kovacs2012multi} and MA-HTN \cite{cardoso2017multi}, respectively, this approach would allow each heterogeneous agent to operate with its own unique pair of HDDL domain and problem files. This capability would enhance HDDLGym's ability to manage complex multi-agent dynamics beyond simple collaboration, supporting scenarios with competition, agent privacy, and distributed context information.

% Future work: \\
% - HDDL-generative system: enable human users to define the domain (hierarchy of doing tasks) themselves with their own thinking process.  \\
% Hierarchical planning results provide strategic insights into the planner’s actions, which can facilitate effective collaboration among multiple agents by enhancing communication. Therefore, HDDLGym offers a promising platform to study interactions between human-human and human-AI agents. As part of our ongoing work, we are developing an HDDL-generative system that interprets human input to construct HDDL domains reflecting individual thought processes. By allowing users to define tasks and actions in a hierarchical structure, we aim to capture human agents' preferences and capabilities in a user-driven way. These customized HDDL domains can then be explored with HDDLGym to produce RL policies tailored to human user preferences. This future work aims to make HDDLGym a powerful tool for developing RL policies that align with human-centered planning and collaborative dynamics in multi-agent environments.
% - LangLTL: convert human description into LTL, then automata 

% Inference system: reverse policy to regenerate the current hierarchy of other agents based on the history of actions and observations
% Transfer learning:
% HDDL → LTL → Automata

\section{Conclusion}
\label{sec:conclusion}
In this work, we introduce HDDLGym, a tool for applying RL to hierarchical planning by converting HDDL-defined problems into Gym environments. Its design balances scalability and functionality, enabling multi-agent interactions and complex task structures. We hope HDDLGym opens new possibilities for studying RL in hierarchical planning, especially in multi-agent contexts.
% In this work, we introduced HDDLGym, a valuable tool for applying RL to hierarchical planning by transforming HDDL-defined problems into Gym environments. Its design balances scalability with functionality, enabling multi-agent interactions and complex task structures. With this, we hope HDDLGym can open new possibilities for studying RL in hierarchical planning, particularly in multi-agents contexts. 
% However, current limitations, such as discrete action spaces, symmetric agent roles, deterministic transitions, and RL policies specific to certain sets of problem files, restrict its broader applicability. Addressing these limitations will expand HDDLGym’s potential for studying RL in complex, real-world, multi-agent scenarios.